\documentclass[journal,twoside,web]{ieeecolor}
\usepackage{generic}
\usepackage{cite}
\usepackage{amsmath,amssymb,amsfonts}
\usepackage{algorithmic}
\usepackage{graphicx}
\usepackage{algorithm,algorithmic}
\usepackage{hyperref}
\hypersetup{hidelinks=true}
\usepackage{textcomp}
\usepackage{booktabs}
\usepackage{hyperref}

\makeatletter
\def\ps@titlepagestyle{%
  \def\@oddhead{}%
  \def\@evenhead{}%
  \def\@oddfoot{}%
  \def\@evenfoot{}%
}
\let\ps@IEEEtitlepagestyle\ps@titlepagestyle

\def\ps@headings{%
  \def\@oddhead{\hbox{}\scriptsize\leftmark \hfil \thepage}%
  \def\@evenhead{\scriptsize\thepage \hfil \leftmark\hbox{}}%
  \def\@oddfoot{}%
  \def\@evenfoot{}%
}
\makeatother

\begin{document}

\title{ReactEMG: Stable, Low-Latency Intent Detection from sEMG via Masked Modeling}
\author{Runsheng Wang$^{1, *}$, Xinyue Zhu$^{2, *}$, Ava Chen$^1$, Jingxi Xu$^{2}$, Lauren Winterbottom$^{3}$, \\Dawn M. Nilsen$^{3, \dagger}$, Joel Stein$^{3, \dagger}$, and Matei Ciocarlie$^{1, \dagger}$
\thanks{$^*$Equal Contribution.}
\thanks{$\dagger$ Co-Principal Investigators.}
\thanks{$^{1}$Department of Mechanical Engineering, Columbia University in the City of New York, NY, USA {\tt\small \{runsheng.w, ava.chen, matei.ciocarlie\}@columbia.edu}}
\thanks{$^{2}$Department of Computer Science, Columbia University in the City of New York, NY, USA {\tt\small xz3013@columbia.edu}, {\tt\small jxu@cs.columbia.edu}}
\thanks{$^{3}$Department of Rehabilitation and Regenerative Medicine, Columbia University Irving Medical Center, New York, NY 10032, USA {\tt\small \{lbw2136, dmn12, js1165\}@cumc.columbia.edu}}
}

\maketitle


\begin{abstract}
Surface electromyography (sEMG) signals show promise for effective human-machine interfaces, particularly in rehabilitation and prosthetics. However, challenges remain in developing systems that respond quickly to user intent, produce stable flicker-free output suitable for device control, and work across different subjects without time-consuming calibration. In this work, we propose a framework for EMG-based intent detection that addresses these challenges. We cast intent detection as per-timestep segmentation of continuous sEMG streams, assigning labels as gestures unfold in real time. We introduce a masked modeling training strategy that aligns muscle activations with their corresponding user intents, enabling rapid onset detection and stable tracking of ongoing gestures. In evaluations against baseline methods, using metrics that capture accuracy, latency and stability for device control, our approach achieves state-of-the-art performance in zero-shot conditions. These results demonstrate its potential for wearable robotics and next-generation prosthetic systems. Our project website, video, code, and dataset are available at: \url{https://reactemg.github.io/}
\end{abstract}



\section{Introduction}
\label{sec:introduction}
    
    \IEEEPARstart{L}{earning-based} methods are increasingly central to human-machine interfaces (HMI), enabling users to control a range of robotic systems through interactive modalities such as wearable sensors. Among these modalities, surface electromyography (sEMG) stands out for its non-invasive ability to detect muscle activity at the skin's surface. Direct sEMG measurement of neuromuscular signals has supported non-clinical applications---including control of multi-fingered grippers~\cite{Loskutova2024, Meeker2022} and drones~\cite{Doshi2021}---as well as clinical uses such as diagnosis of neuromuscular conditions and prosthesis control~\cite{Tchimino2024, reciprocal2025}.
    
\begin{figure*}[t]
    \centering
    \includegraphics[width=0.90\textwidth]{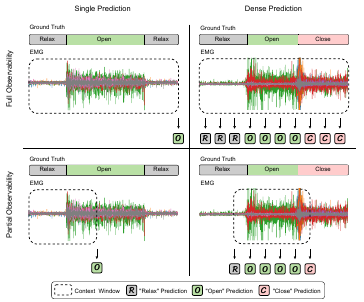}
    \caption{\textbf{Design space for real-time EMG intent detection.} Approaches that rely on full segment observability (top row) wait for an offset signal or a window long enough to cover the entire gesture before making a prediction. They typically assume each gesture is bounded by ``Relax'' or ``Rest'' states, or they segment the region that contains the non-rest gesture beforehand. Since dwell times vary, these methods introduce uncontrolled latency and does not match continuous use. Even dense variants (top‑right) inherit this dependence on known boundaries. With partial observability and a single label (left column), supervision becomes inconsistent when a window straddles a gesture transition---different assignment rules label nearly identical inputs differently---delaying switches and causing flicker that then requires smoothing. Our formulation (bottom‑right) treats intent detection as streaming segmentation: from the observed window it produces dense per‑timestep predictions, enabling immediate prediction and unambiguous supervision at transitions.}
    \label{fig:realtime_paradigm}
    \end{figure*}

    Despite these advances, interpreting EMG for device control remains challenging when models are run online to operate a device continuously in real-time. In this paper, we use the term ``real-time'' to refer to an online setting with two key properties: (i) low \textit{latency}, where the delay from muscle activation to device action is small enough to feel immediate; and (ii) \textit{stability}, where intent predictions remain consistent during sustained ``hold'' periods instead of flickering between classes. This real-time setting differs from offline gesture recognition, which classifies an entire movement after its completion and can thus exploit full temporal observability from gesture onset to offset. In contrast, real-time control must infer intent from partial observations as gestures unfold, handle successive transitions with variable dwell times, and produce per‑timestep outputs suitable for continuous device operation. Prior works have largely emphasized offline evaluation, with relatively few studies reporting online deployment; even when online performance is reported, latency and stability are rarely quantified, and functional task designs often obscure model-specific effects and prevent systematic comparisons across different models. These distinctions are illustrated in~\autoref{fig:realtime_paradigm}.

    More specifically, we use \emph{latency} to refer to the delay from a change in muscle activity to the first correct change in the model's output that would actuate the device. In rehabilitation, low latency matters because building continuous, strong associations between human exertion and observable movement is key for achieving therapeutic outcomes~\cite{Leech2022motorlearning, Micera2020neurotech, Hogan2006engagement} and low latency in closing the human motion loop is needed for therapeutic benefits~\cite{Cisnal2021robhand, Schwartz2016closeloopmovement}. More broadly, HMI applications require low latency because operators have limited patience for input-response delays---wait too long and they will attempt different control inputs to compensate~\cite{Beech2025jitter, Scholcover2018behaviorlatency, Du2024delaycompensation}, leading to unstable robot behaviors~\cite{Neumeier2019teleoperation, Khasawneh2019latency}. 
    
    Once an intent is established, \emph{stability} becomes equally important. We use this term to refer to prediction constancy during holds, with no flicker or spurious class switches. Consider a hand prosthesis assisting a user to grasp a mug: after closing, the system must maintain a ``closed'' prediction throughout the hold. Even brief flicker to ``open'' can cause accidental release, with practical and safety consequences. 
    
    Real-time EMG intent detection therefore requires both rapid, reliable transition detection and robust maintenance of the active intent. The most commonly used metric for such systems---per-time-step accuracy---aggregates performance over an entire window, which can mask critical effects pertaining to these requirements. Therefore, in addition to per‑timestep accuracy, we also focus on response latency and, in order to better capture both latency and stability, we also introduce \textbf{transition accuracy}---a single metric that reflects how quickly the model reacts at transitions and how consistently it maintains the active intent across the hold. Unlike conventional offline metrics, transition accuracy is designed to more accurately and faithfully capture aspects of performance that are most relevant to online device control behaviors.
  
    Beyond real-time performance, another major challenge is zero-shot generalization: a model that works across new users immediately, without subject‑specific calibration or fine‑tuning. This ``plug‑and‑play'' capability is appealing because collecting per‑user EMG data via repetitive gestures is time‑consuming and physically demanding, especially for people with limited mobility. Achieving zero‑shot performance is challenging, however, as EMG varies significantly across individuals and sessions due to anatomical differences, physiological state, and day‑to‑day drift~\cite{DeLuca1997}.
    
    To achieve real-time, stable zero‑shot performance, we present ReactEMG, a combination of a novel learning model, training framework, dataset, and evaluation metric for EMG intent detection. We formulate intent detection as a segmentation problem, in which we predict an action label (i.e., user intent) at each timestep within a continuous EMG stream. For training, we provide both EMG signals and corresponding actions as input, and adopt a masked‑modeling strategy that selectively masks portions of these sequences. By requiring the model to reconstruct the missing segments, we leverage local supervision that anchors muscle activations to user intent, enabling robust signal-action alignment even under imprecise ground‑truth labels. Evaluated with both per-time-step and transition accuracy metrics, our approach achieves state‑of‑the‑art zero‑shot performance on unseen subjects, without any subject‑specific calibration. In summary, our contributions are as follows:

    \begin{itemize}
    
    \item \textbf{State-of-the-art performance for low-latency control.} We target intent prediction for device control that is \emph{responsive} (fast switching and tolerant of rapid successive changes), \emph{stable} (flicker-free holds), and strictly \emph{online} (no hindsight segmentation or offline assumptions). Using transition accuracy, a stringent metric that requires a correct switch within a short window and maintenance of the new intent with zero mislabels, ReactEMG outperforms baselines on EMG-EPN-612 and on our newly collected dataset. Our model also matches or exceeds leading sequence- and convolution-based alternatives on standard per-timestep accuracy and latency metrics.

    \item \textbf{Zero-shot, calibration-free deployment.} ReactEMG is plug-and-play: it requires no subject-specific calibration or fine-tuning and generalizes to unseen users under leave-one-subject-out evaluation. All results are reported in this zero-shot setting using off-the-shelf sEMG hardware. The model maintains robust performance across diverse arm postures, functional grasps, and movement conditions in our dataset, demonstrating immediate practicality for continuous online control without per-user setup.
    
    \item \textbf{Masked multimodal architecture and data‑funnel training.} To achieve the performance described above, we formulate intent detection as streaming segmentation with per‑timestep outputs and a short, bounded look‑ahead for robustness. An encoder‑only transformer jointly encodes EMG and intent tokens and is trained with a dynamic masking strategy that aligns muscle activity with action, improving both onset responsiveness and hold stability. We adopt a data funnel---pretrain on large public EMG datasets, then fine‑tune on a posture‑ and task‑diverse, transition‑rich 28‑subject dataset---achieving higher transition accuracy than training from scratch or pretraining alone.
    \end{itemize}
    

\section{Related Work}
\label{sec:related}

Gesture recognition via sEMG has long been investigated as a control signal for HMI, particularly for prosthetic devices, assistive robotics, and rehabilitation systems~\cite{CHANG1996529,Khokhar2010,Gordleeva2020,Xiong2024,Hargrove2013,Cimolato2022,Ahkami2023, Leonardis2015}. Earlier approaches predominantly relied on extracting handcrafted features from the sEMG signals followed by shallow classifiers such as LDAs~\cite{Antuvan2019} or SVMs~\cite{crawford2005real, Tepe2022, JaramilloYnez2020}. Recent research has shifted toward deep learning paradigms that better exploit sEMG's spatiotemporal structure, including convolutional neural networks (CNNs) for spatial filtering across electrode arrays~\cite{BaronaLpez2024, Betthauser2020}, recurrent networks (RNNs/LSTMs) for modeling temporal dynamics~\cite{Azhiri2021, Simo2019}, and transformer architectures to capture long‑range, cross‑channel interactions via self‑attention~\cite{Zabihi2023, Montazerin2023}.

While strong results are commonly reported on popular datasets (e.g., NinaPro~\cite{Pizzolato2017} and CAPG‑MYO~\cite{Dai2021}), they do not transfer directly to real‑time control. Most evaluations operate on pre‑segmented windows (from labeled onset to offset) or on gesture‑only (non‑relax) excerpts, then aggregate decisions per gesture via majority voting or smoothing---assuming full observability of the entire gesture. This assumption conflicts with continuous control, where observability is partial and gesture end times are unknown. Moreover, methods that predict a single label per window introduce label ambiguity when a window straddles a gesture transition, delaying gesture switches and masking prediction flicker. Figure~\ref{fig:realtime_paradigm} illustrates some of these differences.

Several works aim for real‑time use, make assumptions and use evaluation protocol that present hurdles for real-time integration. Montazerin et al.~\cite{Montazerin2023} report fast recognition with small windows and low latency, but the input is pre‑segmented to exclude the "rest" state, unrealistic for online use. Furthermore, the "fused" variant of their model relies on motor‑unit decomposition performed in hindsight, which the paper acknowledges prevents real‑time deployment. Azhiri et al.~\cite{Azhiri2021} report high performance on a window-level RNN classifiers, but do not quantify streaming properties such as responsiveness of successive transitions or hold‑time flicker, and exclude the "rest" state, limiting online applicability. De Silva et al.~\cite{Silva2020} report high accuracy with fast inference, but predictions are made only at detected onsets and held between onsets, with a 2-second post-prediction refractory period that blocks any subsequent transitions. The pipeline also assumes that flexion is followed by its corresponding extension, which limits applicability to more arbitrary action sequences. 

Reducing the per‑user calibration burden has also been a long‑standing goal in sEMG gesture recognition. Most "subject‑agnostic" approaches still require few‑shot adaptation---via self‑supervision~\cite{Duan2023}, meta‑learning~\cite{LaRotta2024}, generative augmentation~\cite{Xu2025}, or loss design and normalization strategies~\cite{Xue2023}. These methods improve cross‑user robustness but remain calibration‑dependent and are typically validated on segmented windows. Zero‑shot generalization has gained momentum with large, diverse open-source datasets. Eddy et~al.~\cite{Eddy2024} show that training on the EMG‑EPN‑612 dataset improves cross‑user performance under zero-shot settings. However, their protocol (by design) targets discrete gesture recognition that assumes full gesture observability.

In parallel with academia, industry has begun to explore zero‑shot generalization at scale. Meta has released two resources that advance cross-user modeling: \emph{emg2qwerty} (108 users; 346 hours) for typing and \emph{emg2pose} (193 users; 370 hours) for hand pose estimation~\cite{emg2qwerty_neurips2024,salteremg2pose}. Both demonstrate promising zero‑shot results on held‑out users in offline evaluation. While \emph{emg2pose} describes an internal buffered streaming variant of their model, the end‑to‑end setup depends on a proprietary wristband and the model is not open-source, precluding assessment of real‑time latency and stability.

We target the conjunction of real-time operation and zero-shot deployment---an application space that prior work leaves open. On unseen subjects, our model produces dense, per‑timestep labels from only observed data with a small, bounded look‑ahead to stabilize outputs; it does not assume known gesture boundaries and is tuned for fast, reliable switches and stable holds, ideal for real-time continuous online control.


\begin{figure*}[!t]
\centering
\includegraphics[width=\textwidth]{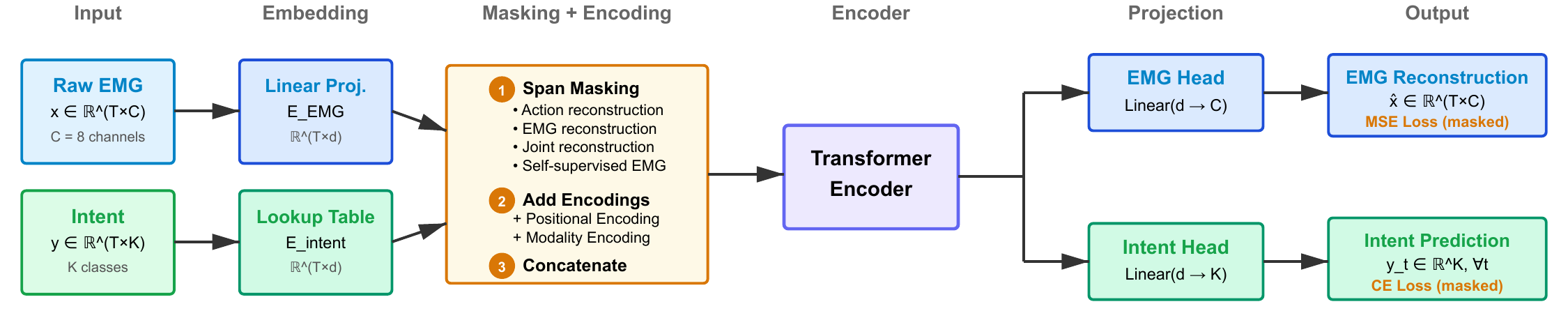}
\caption{\textbf{Model overview}. EMG signals are mapped into an embedding space via a learnable linear projection, while intent tokens use a lookup-based embedding. Both modalities undergo dynamic span masking and receive modality-specific plus shared positional encodings. They are then concatenated and processed by a transformer encoder, after which outputs are split into EMG and intent branches. The EMG branch is optimized via MSE on masked timesteps, and the intent branch via cross-entropy on masked tokens. Losses are added before backpropagation.}
\label{fig:architecture}
\end{figure*}

\section{Method}
\label{sec:method}

Our goal is to continuously predict user intent from forearm muscle activity. We capture an 8-channel EMG signal (Myo armband, Thalmic Labs), sampled at 200~Hz, and train a model to predict the current intent from a predefined set of \(K\) action classes (e.g., ``open hand,'' ``close hand'') at every instant. We formulate this as a segmentation problem over a continuous stream of muscle activity. Let \(x=\{x_1, x_2, \dots, x_T\}\in\mathbb{R}^{T\times C}\) represent a window of length \(T\) of multichannel EMG time series sampled from \(C\) sensors (e.g., eight Myo armband channels). Given \(x\), we define the corresponding per-timestep action likelihood sequence as \(y=\{y_1, y_2, \dots, y_T\}\in\mathbb{R}^{T\times K}\). The goal is to learn a function \( f:\mathbb{R}^{T\times C}\rightarrow\mathbb{R}^{T\times K} \) that returns, for each time index \(t\), a length-\(K\) likelihood vector \(y_t\).

\subsection{Architecture} Although segmentation captures continuous changes in muscle activation, the EMG signal itself remains non-stationary and highly variable across time. Even within a single recording session, different gestures can induce distinct activation patterns, and these patterns may fluctuate due to electrode shift, fatigue, or user variability. Inspired by recent advances in multimodal LLMs and "any-to-any" frameworks~\cite{tang2023anytoany, zhan-etal-2024-anygpt, nextgpt}, we hypothesize that explicitly conditioning EMG representations on the corresponding action label---allowing the model to cross-attend between the two---can mitigate these challenges. To this extent, we propose an encoder-only transformer for EMG and action modeling. Unlike conventional methods that treat EMG signals and their corresponding action labels as an input-output pair, our model treats EMG and intent as two distinct input modalities. We construct a multimodal sequence that includes EMG data followed by the corresponding intent data. The model's objective is to reconstruct the original unmasked sequence---both EMG and intent---at all timesteps. \autoref{fig:architecture} provides an overview of the proposed architecture.

Given a raw EMG signal, we first apply a median filter, then rectify negative amplitudes, and extract overlapping sliding windows. Our model treats EMG signals and subject intent as two separate modalities, each mapped to its own space. For EMG, we use a learnable linear projection to map the raw 8-channel signal directly into the embedding space, without any manual feature extraction. In parallel, we represent the subject's intent in the embedding space via a standard lookup table (covering valid intents plus a dedicated mask token).

Next, we apply span masking to the projected EMG and intent embeddings. Following a protocol similar to RoBERTa's~\cite{zhuang-etal-2021-robustly} dynamic masking, we randomly mask out contiguous timesteps in either or both modalities and regenerate these masks at each training epoch, thereby exposing different regions of the same sequence to the model over time. After masking, we add modality-specific encoding vectors and positional encodings to both the EMG and action tokens, ensuring each token can be distinguished not only by its position in time but also by the modality it belongs to. These positional encodings are shared across both modalities to enforce consistent temporal alignment.

We then concatenate the EMG and intent sequences into a single multimodal sequence and feed it into a transformer encoder to learn cross-modal dependencies. Specifically, let \(\tilde{x}=\mathrm{Enc}_{\text{emg}}(x)\) and \(\tilde{y}=\mathrm{Enc}_{\text{int}}(y)\) denote the masked, position- and modality-encoded embeddings; we then form the multimodal sequence \(z=[\tilde{x};\,\tilde{y}]\) by concatenating along the sequence dimension, where \([\cdot;\cdot]\) denotes concatenation. Following the transformer encoder, we split the output multimodal sequence back into EMG and intent constituents. The EMG output is projected through a linear layer to produce eight values per timestep, and the loss on EMG is computed against raw 8-channel values using mean-squared error (MSE) over only the masked positions. The intent output goes through a separate linear layer for dense classification at every timestep, with a cross-entropy loss on masked intent tokens.

\subsection{Masking strategy}
\label{sec:masking_strategy}

We train with a multimodal masked-reconstruction objective that ties EMG and intent temporally. Using earlier notation, let $x\in\mathbb{R}^{T\times C}$ be the EMG window and $y\in[K]^T$ the corresponding per‑timestep action sequence. During training, the model consumes \emph{masked} EMG and \emph{masked} intent tokens and is asked to reconstruct both; at inference, only EMG is provided and all intent tokens are masked.

\noindent\textbf{Interface and notation.}
We maintain an EMG mask set $\mathcal{M}_{E}$ and an action mask set $\mathcal{M}_{A}$:
\[
\mathcal{M}_{E}\subseteq\{1,\dots,T\}\times\{1,\dots,C\},\qquad
\mathcal{M}_{A}\subseteq\{1,\dots,T\}.
\]
Masked EMG samples are embedded by swapping their embeddings with a learned vector $\mathbf{m}\in\mathbb{R}^{d}$, while masked intent tokens are replaced with a dedicated mask token $\langle\!\text{MASK}\!\rangle$. Thus the model receives:
\[
\begin{aligned}
\tilde{x}_{t,c} &=
\begin{cases}
\mathbf{m}, & (t,c)\in\mathcal{M}_{E},\\
x_{t,c}, & \text{otherwise},
\end{cases}
\\[2pt]
\tilde{y}_{t} &=
\begin{cases}
\langle\!\text{MASK}\!\rangle, & t\in\mathcal{M}_{A},\\
y_{t}, & \text{otherwise}.
\end{cases}
\end{aligned}
\]

\noindent\textbf{Training tasks.}
We define the following masking tasks, which together span the regimes required for real-time control. Each sample is duplicated once per task so it participates in all tasks, improving sample efficiency:
\begin{itemize}
\item \emph{Action reconstruction (primary intent task).} Mask a subset of intent timesteps ($\mathcal{M}_{A}\neq\varnothing$, $\mathcal{M}_{E}=\varnothing$) and predict the missing actions from unmasked EMG and surrounding intent context. This anchors EMG patterns to intent without requiring entire gestures to be visible.
\item \emph{EMG reconstruction.} Mask a subset of EMG timesteps ($\mathcal{M}_{E}\neq\varnothing$, $\mathcal{M}_{A}=\varnothing$) and reconstruct the missing EMG conditioned on visible intent. This stabilizes the shared representation and improves cross‑modal alignment.
\item \emph{Joint reconstruction.} Mask aligned spans in both modalities ($\mathcal{M}_{E}\neq\varnothing$ and $\mathcal{M}_{A}\neq\varnothing$) at the same timesteps to force temporal coupling between muscle activity and intent, improving sensitivity at transitions and robustness during holds.
\item \emph{Self‑supervised EMG (unlabeled).} When given unlabeled data with missing intent labels, we block attention from EMG to intent via an attention mask and replace intent tokens with the $\langle\!\text{MASK}\!\rangle$ token. The model reconstructs EMG from EMG context alone.
\end{itemize}

\noindent\textbf{Mask generation.}
Masks are created by sampling a mask type and then generating time indices:
\begin{itemize}
\item \emph{Span‑based masking (channel‑aligned).} Draw contiguous time spans from a Poisson distribution with scale parameter $\lambda$. For each selected timestep $t$ in a span $S$, EMG masking covers all channels, i.e., $(t,c)\in\mathcal{M}_{E}$ for every $c$. This yields realistic, temporally coherent occlusions.
\item \emph{End‑of‑window masking.} Select a suffix of the window and mask that interval. This simulates partial observability near gesture offsets and enforces responsiveness to onsets without relying on known boundaries. Note that sliding windows are themselves arbitrary: they likewise make no boundary assumptions, and no observability criterion defines a window---it simply slides over the data.
\item \emph{Targeted transition masking (intent).} For windows containing a change point $\tau$ in $y$, mask intent within a temporal buffer around $\tau$ while leaving EMG unmasked. This focuses learning on fine-grained gesture transition behaviors that matters for real‑time use.
\end{itemize}
The mask type for each training instance is sampled from a task‑dependent mixture with coefficients $\{\pi_{\text{span}},\pi_{\text{end}},\pi_{\text{transition}}\}$ (the transition option applies only to the action reconstruction task). For each selected mask type, a masking proportion $p_{t}\in[p_{\min}^{(t)},\,p_{\max}^{(t)}]$ determines the number of timesteps to mask, $N_{t}=\lfloor p_{t}\,T\rfloor$; spans are sampled and concatenated until at least $N_t$ timesteps are covered. For unlabeled windows, only span‑based EMG masking is used. Mask parameters are re‑sampled every epoch to expose diverse occlusion patterns over the same data.

\noindent\textbf{Objective.}
Let $f_{A}$ and $f_{E}$ denote the intent and EMG reconstruction heads. Losses are computed only on masked positions:
\[
\begin{aligned}
\mathcal{L}
&=
\frac{1}{|\mathcal{M}_{E}|}
\sum_{\substack{(t,c)\in\mathcal{M}_{E}}}
\bigl\| f_{E}(\tilde{x},\tilde{y})_{t,c}-x_{t,c}\bigr\|_2^{2}
\\
&\quad+\frac{1}{|\mathcal{M}_{A}|}
\sum_{\substack{t\in\mathcal{M}_{A}}}
\mathrm{CE}\!\bigl( f_{A}(\tilde{x},\tilde{y})_{t},\,y_{t} \bigr).
\end{aligned}
\]

If one mask set is empty, its term is omitted. At inference, we set $\mathcal{M}_{A}=\{1,\dots,T\}$ and $\mathcal{M}_{E}=\varnothing$ (i.e., all intent tokens are masked, EMG is unmasked), so the network outputs per‑timestep class scores $\hat{y}=f_{A}(x)$ for real‑time control.

\subsection{Real-Time Inference}
\label{sec:online_inference}

Real-time inference does not come for free. Unlike offline evaluation, an online system must decide from a live stream under tight latency and compute budgets, with only partial observability and no oracle boundaries. Consequently, design choices and pipeline assumptions that are harmless offline quickly become untenable and unrealistic at run time: requiring known onset/offset boundaries (full-segment observability)~\cite{Eddy2024, Montazerin2023}, removing the \emph{rest/relax} state~\cite{Montazerin2023}, relying on preprocessing statistics computed offline~\cite{Azhiri2021}, producing a single label per window even when it straddles a transition~\cite{Lee2021, Zabihi2023}, or imposing action‑sequence rules~\cite{Silva2020}. 

There are many valid ways to build a real-time inference framework; the key is to make the latency-stability trade-offs explicit and tune them to the task. During online deployment, our trained model processes a continuous stream of EMG windows and generates a label for each timestep in every window. While we could assign a label to a timestep as soon as it appears in the sliding window, transient noise or incomplete gestures often lead to flickering. Since our model produces dense predictions and the windows overlap, multiple predictions are available for the same timestep. We can fuse these overlapping predictions into a single label through an aggregator that explicitly controls for latency and stability. Here, we present a smoothing method that is applicable to all models with dense output.

\begin{figure*}[t]
\centering
\includegraphics[width=0.8\textwidth]{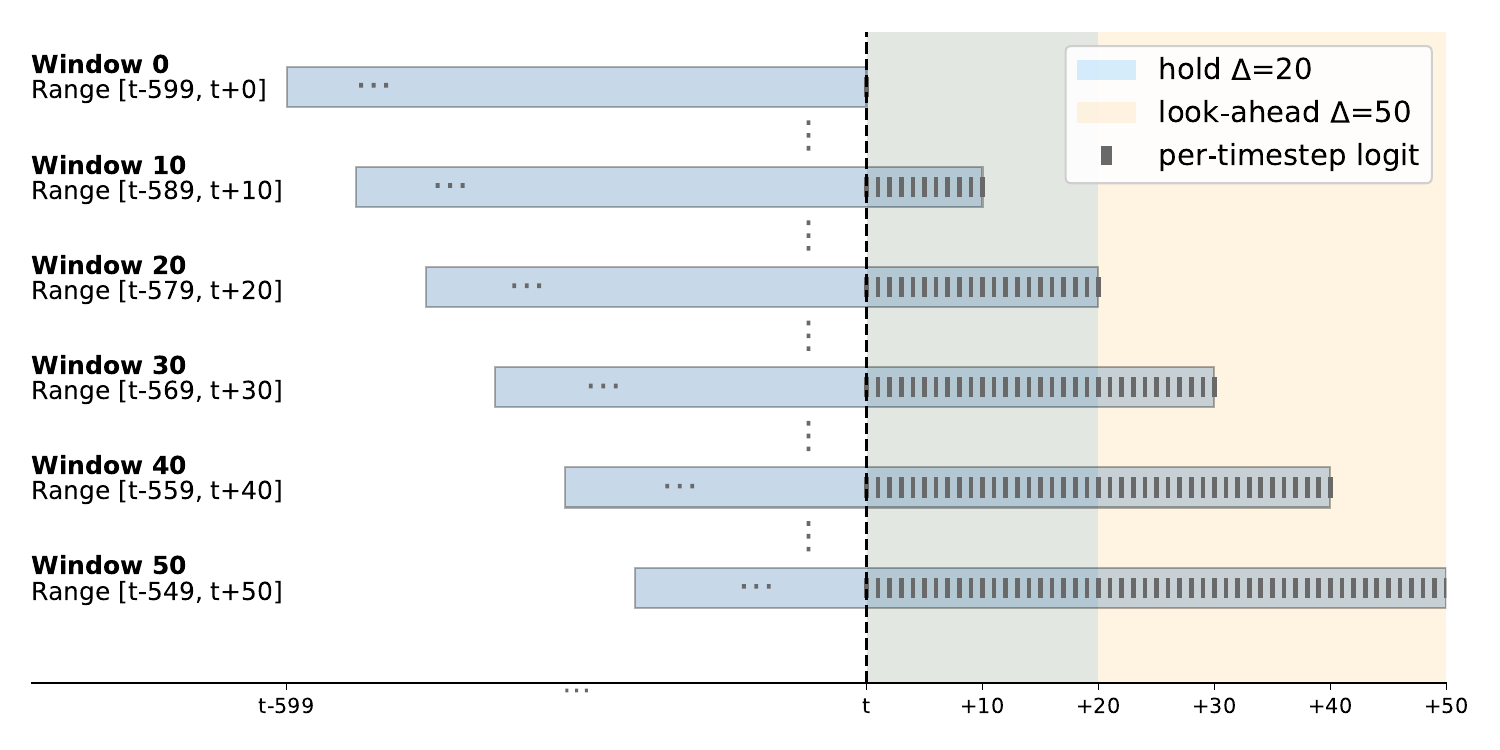}
\caption{\textbf{Online inference with look-ahead horizon and reduced inference frequency}. Five representative overlapping sliding windows are depicted, ending at time steps $t,t{+}10,\dots,t{+}50$ respectively. The small grey vertical bars within each window indicate the per-timestep logits that make up the average.}
\label{fig:smoothing}
\end{figure*}

Suppose we want to predict the intent at timestep $t$. As shown in \autoref{fig:smoothing}, we evaluate all overlapping windows of length \(T\) (sampled at \(f_s = 200\,\mathrm{Hz}\)) whose final sample lies in \([t,\, t{+}\ell]\), where \(\ell\) is the look-ahead in timesteps. For each window, the model emits a logit at every timestep; we aggregate the logits over the look-ahead horizon and apply \(\arg\max\) to obtain the label at \(t\). To suppress jitter, we hold the emitted label for \(s\) timesteps, then repeat at \(t{+}s\). The resulting update frequency is \(f_s/s\). For the smoothed variant of our model, we use a window size of \(T=600\), a look-ahead of \(\ell=50\) timesteps (\(0.25\,\mathrm{s}\)), and a label hold time of \(s=20\) timesteps (\(0.1\,\mathrm{s}\)), yielding an effective update rate of \(10\,\mathrm{Hz}\).

Our approach introduces two sources of latency. First, the label at time $t$ is issued after ingesting data up to $t+\ell$, introducing a delay of up to $\ell$ in responding to changes in intent. This delay is smaller than typical human reaction time to visual or verbal stimuli~\cite{Woods2015, Jain2015, Arafat2022Comparison}. Second, the hold period prevents the model from reacting to intent changes faster than 10~Hz. We consider it highly unlikely that a human subject would modulate hand pose or intent at such a rate.

We note that any method that uses future samples introduces delay. In our case, the delay is bounded by $\ell/f_s$, which does not by itself preclude real‑time operation. Our formulation avoids the unrealistic assumptions discussed earlier; its smoothing is limited to short‑horizon aggregation within $[t,\,t{+}\ell]$ plus a brief hold period $s$. Additionally, our model operates directly in the raw signal space, applying only minimal preprocessing that imposes no data format assumptions and keeps the pipeline fast to run and easy to deploy online.

\subsection{Datasets}
\label{sec:dataset_description}

Capturing the diversity of EMG signals is essential for robust zero-shot generalization. To this end, we aggregate four open‑source sEMG datasets~\cite{epn_dataset, roshambo_dataset, Kanoga2021, mangalore_dataset} recorded with a single Myo armband at 200~Hz and whose gestures can be mapped to a common six‑gesture set: \emph{hand relax}, \emph{hand open}, \emph{hand close},
\emph{wave-in}, \emph{wave-out}, and \emph{pinch}. We do not include other datasets so that sampling rate, sensor configuration, and gesture taxonomy remain compatible with our setting. 

~\autoref{tab:publicEMG} provides an overview of the public datasets used. The recording protocols for each dataset are as follows:

\begin{itemize}
  \item \emph{EMG-EPN-612}: Each recording is \(\sim\)5\,s; a single gesture can be performed at any time within the window; no strictly enforced maintenance period.
  \item \emph{ROSHAMBO}: Each recording includes the gestures rock, paper, and scissors, each performed five times and held for 2\,s, with \emph{relax} intervals between gestures.
  \item \emph{SS-STM}: Each \(\sim\)5\,s recording: 1\,s preparatory \emph{relax}, 2\,s gesture hold, then 2\,s \emph{relax}.
  \item \emph{Mangalore}: Each recording is a single gesture repeated 25 times; each repetition has a 2\,s hold followed by 2\,s \emph{relax}.
\end{itemize}

\begin{table}[h]
\centering
\caption{Overview of public EMG datasets used in this study}
\resizebox{\columnwidth}{!}{%
\begin{tabular}{lcccc}
\toprule
\textbf{Dataset} & \textbf{\# Subjects} & \textbf{\# Gestures} & \textbf{Gesture Duration} & \textbf{Repetitions} \\
\midrule
\textbf{EMG-EPN-612} & 612 & 6 & 5\,s mixed & 50 \\
\textbf{ROSHAMBO} & 10 & 3 & 2\,s hold + 1\,s rest & 5 \\
\textbf{SS-STM} & 25 & 22 & 1\,s prep + 2\,s hold + 2\,s rest & 5 \\
\textbf{Mangalore } & 10 & 5 & 2\,s hold + 2\,s rest & 100 \\
\bottomrule
\end{tabular}%
}
\label{tab:publicEMG}
\end{table}

While these datasets are valuable, they have limitations for continuous control: gestures are typically maintained for at most 2\,s, there are no transitions between non-\emph{relax} gestures (e.g., open to close), and variability in arm position and grasp type (e.g., cylindrical vs.\ spherical) is limited. To address these gaps, we collected \textbf{ROAM-EMG}, a 28-subject dataset spanning a broader set of real-world conditions, including varied arm positions, functional grasps, and explicit transitions between non-\emph{relax} gestures. All participants provided informed consent to procedures approved by the Columbia University Institutional Review Board (Protocol \#AAAS8104).

Data were collected with a Myo armband, a stress ball, a Campbell soup can, and a soft tape measure. The LED electrode was aligned with the middle-finger metacarpophalangeal (MCP) joint of the right hand. We measured distances from the LED electrode to the ulnar styloid and olecranon and recorded forearm circumference. On slimmer forearms, we wrap an elastic hair tie or rubber band around the Myo for improved electrode-skin contact. All sets were recorded with participants seated.

The protocol began with static-pose data collection using the sequence "ROCORORCR," where \textbf{R} denotes \emph{relax}, \textbf{O} denotes \emph{open}, and \textbf{C} denotes \emph{close}, each held for five seconds. Our dataset targets three functional intents---\emph{open}, \emph{close}, and \emph{relax}---chosen for their relevance to hand exoskeletons and parallel grippers, and sequences include non-\emph{relax} to non-\emph{relax} transitions by design. We recorded four static sets: Set~0 (\emph{Static Relaxing}) with elbow aligned to the table edge, forearm resting on the table, and the base of the palm fully supported; Set~1 (\emph{Static Hanging}) with the arm hanging perpendicular to the ground; Set~2 (\emph{Static Unsupported}) with the arm unsupported at ~45$^\circ$ from the body and a 90$^\circ$ elbow bend; and Set~3 (\emph{Static Reaching}) with the arm fully extended forward without support. We then recorded two functional grasping sets: Set~4 (stress ball) and Set~5 (soup can), in which participants repeatedly grasped and released the object with medium effort. Finally, Set~6 (\emph{Movement}) captured sustained \emph{relax}, \emph{open}, and \emph{close} while participants traced horizontal and vertical circles in both clockwise and counterclockwise directions.

\subsection{Training}
\label{sec:training_description}

We train our model in two setups. First, on the \emph{EMG-EPN-612} benchmark, we train exclusively on its training split while holding out the test set, reporting results for both the full six-gesture model and a three-gesture subset. Second, to assess zero-shot generalization under posture and task variation, we pretrain on all available public datasets and then fine-tune on our dataset under a leave-one-subject-out protocol, withholding each test subject's data from training. In both cases, we retain the checkpoint with the lowest validation loss. ~\autoref{tab:hyperparams} lists all the hyperparameters and training configurations used for our model.

\begin{table}[h]
\centering
\caption{Model hyperparameters and training configuration for our model}
\label{tab:hyperparams}
\begin{tabular}{l l}
\toprule
\textbf{Parameter} & \textbf{Value} \\
\midrule
Seed & 42 \\
Window size & 600 \\
Median filter size & 3 \\
Sliding window stride & 30 \\
Embedding dimension & 128 \\
Number of attention heads & 4 \\
Dropout & 0.15 \\
Activation & GELU \\
Number of encoder layers & 2 \\
Batch size & 128 \\
Epochs & 12 \\
Learning rate & 1e-4 \\
Warmup ratio & 0.05 \\
Decay method & Linear \\
Optimizer & AdamW \\
$\lambda$ (Poisson span masking) & 7 \\
\bottomrule
\end{tabular}%
\end{table}


\begin{table*}[t]
\centering
\caption{Performance on the EMG-EPN-612 (3-class and 6-class) and our ROAM-EMG datasets. Values are mean~$\pm$~SD.}
\label{tab:main_result}
\small
\setlength{\tabcolsep}{6pt}
\begin{tabular}{l c cc cc cc}
\toprule
\multicolumn{2}{c}{} &
\multicolumn{2}{c}{\textbf{EPN 3-Class}} &
\multicolumn{2}{c}{\textbf{EPN 6-Class}} &
\multicolumn{2}{c}{\textbf{ROAM-EMG}} \\
\cmidrule(lr){3-4}\cmidrule(lr){5-6}\cmidrule(l){7-8}
\textbf{Method} & \textbf{Lookahead} &
\textbf{Raw Acc.} & \textbf{Trans. Acc.} &
\textbf{Raw Acc.} & \textbf{Trans. Acc.} &
\textbf{Raw Acc.} & \textbf{Trans. Acc.} \\
\midrule
LDA          & 0\,s      & $0.74 \pm 0.04$ & $0.17 \pm 0.08$ & $0.72 \pm 0.05$ & $0.10 \pm 0.06$ & $0.76 \pm 0.05$ & $0.10 \pm 0.07$ \\
MLP          & 0\,s      & $0.86 \pm 0.04$ & $0.61 \pm 0.12$ & $0.81 \pm 0.04$ & $0.42 \pm 0.11$ & $0.70 \pm 0.07$ & $0.14 \pm 0.11$ \\
LSTM         & 0\,s      & $0.90 \pm 0.03$ & $0.80 \pm 0.12$ & $0.86 \pm 0.03$ & $0.66 \pm 0.14$ & $0.80 \pm 0.08$ & $0.25 \pm 0.17$ \\
ED-TCN       & 0\,s      & $0.94 \pm 0.03$ & $\mathbf{0.84 \pm 0.11}$ & $0.90 \pm 0.03$ & $0.72 \pm 0.12$ & $0.83 \pm 0.07$ & $0.30 \pm 0.15$ \\
TraHGR       & 0\,s      & $0.94 \pm 0.03$ & $0.61 \pm 0.12$ & $0.90 \pm 0.04$ & $0.41 \pm 0.10$ & $\mathbf{0.85 \pm 0.06}$ & $0.33 \pm 0.17$ \\
\textbf{Ours}& 0\,s      & $\mathbf{0.95 \pm 0.03}$ & $\mathbf{0.84 \pm 0.10}$ & $\mathbf{0.92 \pm 0.03}$ & $\mathbf{0.73 \pm 0.12}$ & $\mathbf{0.85 \pm 0.06}$ & $\mathbf{0.36 \pm 0.18}$ \\
\addlinespace[3pt]
LSTM         & 0.25\,s   & $0.92 \pm 0.03$ & $0.84 \pm 0.12$ & $0.88 \pm 0.04$ & $0.71 \pm 0.13$ & $0.81 \pm 0.08$ & $0.32 \pm 0.18$ \\
ED-TCN       & 0.25\,s   & $\mathbf{0.95 \pm 0.03}$ & $\mathbf{0.86 \pm 0.11}$ & $\mathbf{0.92 \pm 0.04}$ & $0.74 \pm 0.13$ & $0.85 \pm 0.07$ & $0.42 \pm 0.18$ \\
\textbf{Ours}& 0.25\,s   & $\mathbf{0.95 \pm 0.03}$ & $\mathbf{0.86 \pm 0.11}$ & $\mathbf{0.92 \pm 0.03}$ & $\mathbf{0.76 \pm 0.13}$ & $\mathbf{0.87 \pm 0.07}$ & $\mathbf{0.52 \pm 0.18}$ \\
\midrule
\multicolumn{2}{l}{\emph{Number of Subjects}} & 306 & 306 & 306 & 306 & 28 & 28 \\
\bottomrule
\end{tabular}
\end{table*}

\section{Evaluation}
\label{sec:results}

\subsection{Evaluation metrics}
Many prior studies report window-level or segment-level classification accuracy, a metric that does not capture performance in continuous online settings. Approaches that aim to approximate real-time use typically assume unrealistic online conditions and rarely report latency or stability~\cite{BaronaLpez2024}. Online user testing can reveal these effects; however, small sample sizes and substantial human variability make it difficult to control confounding factors and attribute observed differences to the model architecture. To standardize model evaluation in offline for real-time device control, we report three complementary metrics: \emph{raw accuracy}, \emph{transition accuracy}, and \emph{latency}.

\textbf{Raw Accuracy.} raw accuracy measures how often the predicted label at each EMG timestep matches the ground-truth label, serving as a straightforward proxy for overall performance in continuous control scenarios. Since our approach generates predictions at every EMG sample, it inherently supports per-timestep computations. For models that produce labels at coarser intervals, we upsample their outputs to the raw EMG sampling rate to enable a fair, per-timestep comparison under realistic online conditions. 

\textbf{Transition Accuracy.} Raw accuracy hides the temporal structure of errors: delay and flicker both register as "misclassification," and heavy imbalance between \emph{``relax"} and the other gesture classes makes those errors look rare. As a result, a system that achieves high raw accuracy can feel unusably slow or flicker uncontrollably. We observe that in real-time online EMG inference---especially for device control---the EMG signal can be separated into two temporal components: transition, the interval when muscle activity shifts between gestures; and maintenance, the interval when the current gesture is sustained. To capture both aspects in evaluation, we introduce \emph{transition accuracy}: a real-time metric designed to faithfully capture latency and stability in online use. 

\begin{figure}[t]
\centering
\includegraphics[width=\columnwidth]{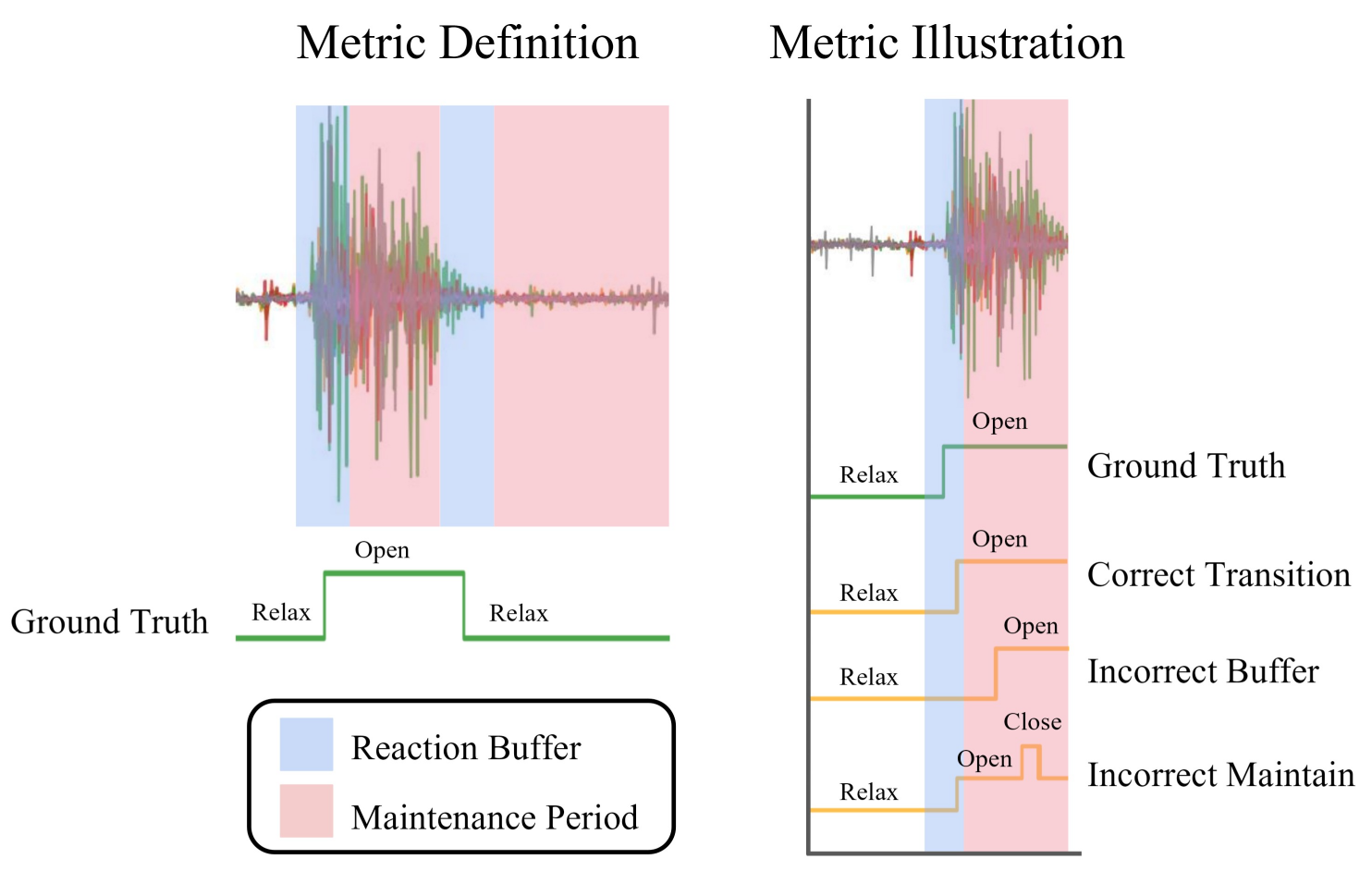}
\caption{\textbf{Transition accuracy Evaluation Protocol}. For each ground‐truth transition from class \(y_{\mathrm{old}}\) to \(y_{\mathrm{new}}\), we define a \emph{reaction buffer} (blue region) centered around the ground truth transition. The model must predict \(y_{\mathrm{new}}\) at least once within this buffer and must not predict any other class. The subsequent \emph{maintenance period} (red region) extends from the end of the reaction buffer to the start of the next reaction buffer. During this interval, the model must predict \(y_{\mathrm{new}}\) at all times. A transition is scored as correct only if both reaction buffer and maintenance period conditions are satisfied.}
\label{fig:transition}
\end{figure}

We define a transition as any change in the user's intent label from \(y_{\mathrm{old}}\) to \(y_{\mathrm{new}}\). As illustrated in~\autoref{fig:transition}, each transition is divided into two disjoint intervals:
\begin{itemize}
    \item Reaction buffer: A short time window (typically 0.5\,s) centered around the ground truth intent transition timestep. The model must predict transition from \(y_{\mathrm{old}}\) to \(y_{\mathrm{new}}\) at least once within this buffer, and should not contain any other classes. We use this buffer window to account for the fact that gesture labels rarely align with the true onset time. The delay between an experimenter/computer program issuing a cue to the subject executing the corresponding gesture is unknown. Subjects may also anticipate cues, activating muscles before the ground truth label shift. Manually correcting these misalignments via relabeling introduces bias (e.g., using an arbitrary threshold, or visual inspection), and is impractical for large datasets, such as the EMG-EPN-612.
    \item Maintenance period: The maintenance period spans the remainder of the gesture until the onset of the next reaction buffer. After predicting \(y_{\mathrm{new}}\), the model must consistently output the same label for the entire duration of this interval. Any incorrect prediction within this interval immediately invalidates the entire transition. Prediction on a transition is considered correct if and only if both these conditions are met. By penalizing both delayed onset detection and instability during the maintenance period, this metric provides a stringent and realistic measure of online performance.
\end{itemize}

\textbf{Latency.} To quantify latency more directly, we measure the offset between each predicted transition and its ground truth counterpart, restricting the analysis to transitions counted as correct under the transition accuracy metric. Since the reaction buffer is centered on the annotated transition, offsets may be positive or negative. We aggregate early and late offsets by magnitude and emphasize that the "ground-truth" annotation is a noisy, imprecise proxy for the true onset.

In all evaluations, we simulate the online streaming setup offline by feeding the model window-by-window in chronological order, as if samples arrive sequentially, and generate outputs using the setup described in Section~\ref{sec:online_inference}.

\subsection{Baselines}

We compare against baselines that are both representative of and competitive within the sEMG literature. We select one well-established method from each architecture family: (i) shallow classifiers; (ii) feature-based feed-forward networks; (iii) recurrent sequence models; (iv) temporal convolutional networks; and (v) transformers. Since our study targets real-time behaviors, published results are not directly comparable to our streaming setting, and public code is often unavailable. Therefore, we evaluate all baselines under a unified pipeline with the same data splits on our transition accuracy metric. Where official implementations are missing, we provide reimplementations, following the authors' stated preprocessing and training details and changing only what is necessary to align with our setting. All baseline methods were carefully tuned to achieve strong performance on our task. Concretely, we benchmark five representative baselines---LDA, MLP, LSTM, ED-TCN, and TraHGR---summarized below.

\textbf{LDA}~\cite{Qi2019-xt} uses a linear discriminant classifier over hand-crafted features computed on sliding windows. The featurization extracts a compact set of discriminative metrics (e.g., RMS, waveform length, and median amplitude spectrum) per channel, standardize them, then apply a linear discriminant projection. Whereas Qi et al.\ pair their feature transform with a GA-optimized ELM classifier, we instead use the discriminant itself for classification to maintain a simple shallow baseline that produces one label per window.

\textbf{MLP}~\cite{Lee2021} employs a sliding-window strategy and assigns a single label to each window. The preprocessing stage extracts a compact set of time-domain metrics---such as RMS, variance, mean absolute value, and zero crossings---from each channel in the window. After standardization, these features are fed into a multi-layer perceptron, which outputs one predicted gesture label per window.

\textbf{LSTM}~\cite{Simo2019} processes EMG as a sequence, using fixed-length segments to train a time-distributed classifier. A fully connected layer first maps the per-timestep inputs to a higher-dimensional representation, which is then fed to an LSTM layer that captures temporal dependencies. A final layer produces gesture class scores at each timestep, yielding dense predictions that are causal within the segment.

\textbf{ED-TCN}~\cite{Betthauser2020} is an encoder-decoder temporal convolutional network that relies solely on 1D convolutions. The encoder applies progressively downsampled temporal convolutions with relatively long kernels and temporal max-pooling, while the decoder mirrors this process through upsampling and additional convolutional layers. A final time-distributed dense layer projects the decoder outputs into the target class dimension, producing per-timestep gesture classifications.

\textbf{TraHGR}~\cite{Zabihi2023} operates on the raw signal space via patch embeddings. Each window is split into non-overlapping patches along two paths---temporal (per-channel) and featural (across channels)---which are linearly projected to tokens, augmented with learned positional embeddings, and prepended with a class token. The token sequences pass through Transformer encoder blocks; the class-token representations are fused and fed to a linear head to produce one label per window. The model is trained fully supervised with cross-entropy.

\subsection{Results and ablations}

As shown in \autoref{tab:main_result}, our approach achieves the best performance on both metrics across all experimental settings. Baseline methods degrade when the evaluation shifts from raw to transition accuracy, whereas our method remains robust. The gap widens under zero-shot evaluation on our dataset, which features more varied conditions. These results indicate that our masked modeling segmentation framework is well suited to zero-shot, real-time intent detection, enabling fast onset detection and stable gesture maintenance. We report results for look-ahead windows of 0~s (no lookahead) and 0.25~s; the window size sets a lower bound on end-to-end latency (see~\autoref{sec:online_inference}). We omit smoothed results for single-label architectures because their single-label nature precludes the use of our smoothing method. \autoref{tab:latency_singlecol} compares the prediction offset of our model against the best baseline method. Our method is able to achieve lower average and median prediction offsets than ED-TCN on both the 3-class and 6-class EPN settings. A video demonstration of online zero-shot inference is available on our \href{https://reactemg.github.io/}{project website}.

\begin{table}[t]
\centering
\caption{Prediction offset on EPN (3- and 6-class). Values are in milliseconds. Lower is better.}
\label{tab:latency_singlecol}
\footnotesize
\setlength{\tabcolsep}{4pt}
\begin{tabular}{@{}lcccc@{}}
\toprule
& \multicolumn{2}{c}{\textbf{EPN 3-Class}} &
  \multicolumn{2}{c}{\textbf{EPN 6-Class}} \\
\cmidrule(lr){2-3}\cmidrule(lr){4-5}
\textbf{Method} & \textbf{Avg. Offset} & \textbf{Med. Offset} &
\textbf{Avg. Offset} & \textbf{Med. Offset} \\
\midrule
ED-TCN & 154 & 130 & 157 & 130 \\
\textbf{Ours} & \textbf{120} & \textbf{85} & \textbf{122} & \textbf{85} \\
\bottomrule
\end{tabular}
\end{table}

\textbf{Pretraining.} We conduct ablations to quantify the effect of pretraining (see \autoref{tab:roam_ablation}). We compare three settings: (1) pretrain-then-fine-tune, (2) training from scratch (no pretraining), and (3) pretraining only. Aggregating multiple open-source datasets broadens subject coverage and encourages subject-invariant representations, but on its own underperforms in real-world settings because these datasets do not capture diverse conditions (e.g., arm poses and object interactions). To combine breadth with robustness, we pretrain on the large open-source pool and then fine-tune on a smaller, targeted dataset with varied poses and interactions. This pretrain-then-fine-tune approach yields the highest transition accuracy, indicating that pretraining improves data efficiency when adapting to a more diverse distribution. In contrast, evaluating the pretrained-only model directly on our dataset performs poorly, consistent with a distribution mismatch: public datasets have relatively narrow, fixed recording conditions, whereas our dataset spans a broader range.

\begin{table}[h]
  \centering
  \caption{Pretraining Ablations on Our Dataset.}
  \label{tab:roam_ablation}
    \begin{tabular}{l c c c}
      \toprule
      \textbf{Method} & \textbf{Lookahead} & \textbf{Raw Acc.} & \textbf{Transition Acc.} \\
      \midrule
      No Fine-tuning & 0s & 0.61 & 0.02 \\
      No Pretraining & 0s & 0.68 & 0.03 \\
      Pretrain + Fine-tune & 0s & \textbf{0.85} & \textbf{0.36} \\
      \midrule
      No Fine-tuning & 0.25s & 0.61 & 0.01 \\
      No Pretraining & 0.25s & 0.70 & 0.06 \\
      Pretrain + Finetune & 0.25s & \textbf{0.87} & \textbf{0.52} \\
      \bottomrule
    \end{tabular}%
\end{table}

\begin{figure}[h]
\centering
\includegraphics[width=1.0\columnwidth]{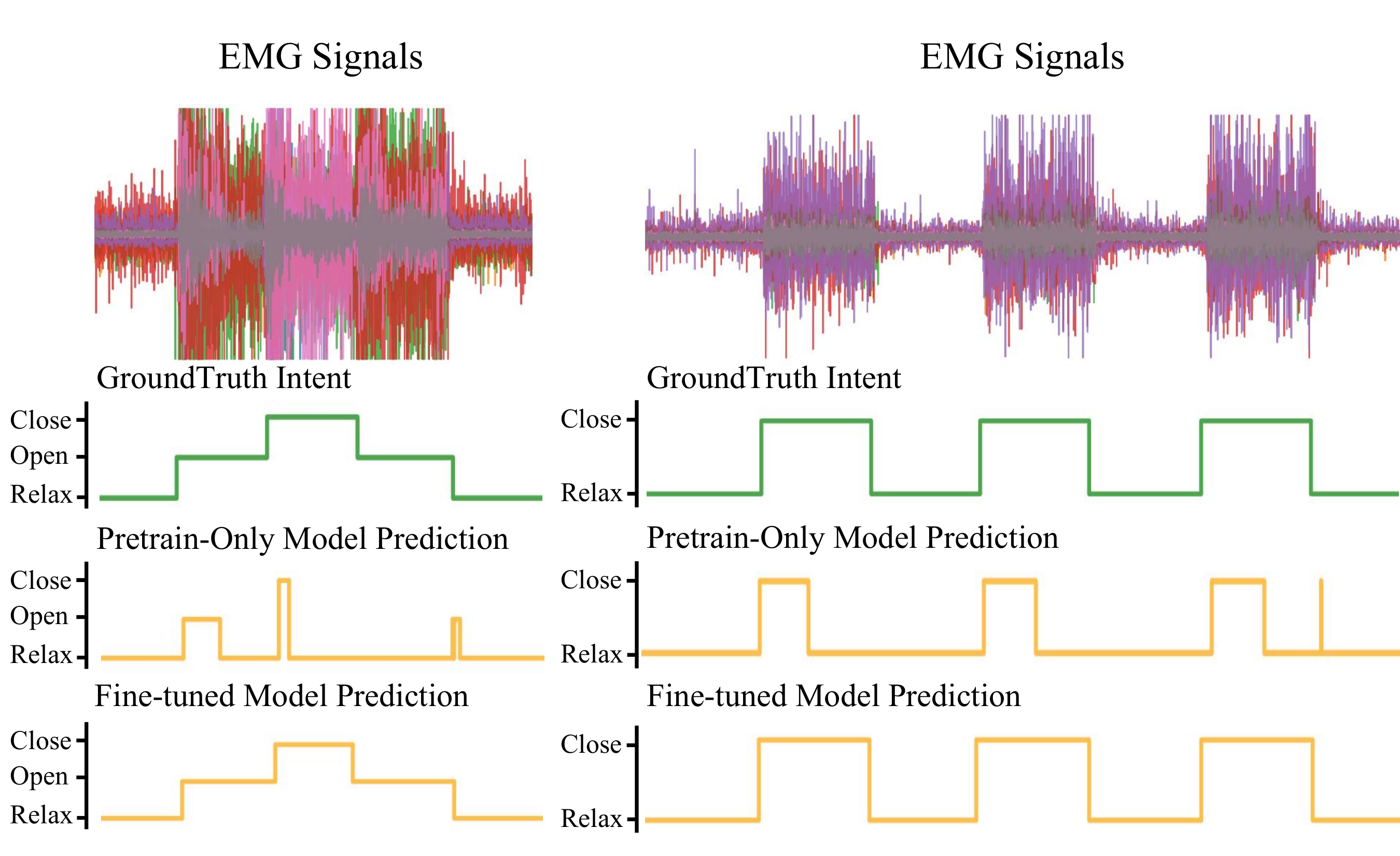}
\caption{Qualitative effect of fine-tuning on our dataset (ROAM-EMG). The public-data-only model produces brief, flickering gesture predictions that quickly return to \textit{rest}---capturing onsets but not maintenance---whereas after fine-tuning the same architecture tracks both onset and sustained gestures and handles transitions that bypass \textit{rest}.}
\label{fig:epnvsroam}
\end{figure}

The effect of fine-tuning after public data pretraining is also evident in the outputs. As shown in \autoref{fig:epnvsroam}, a model trained only on public datasets produces short, flickering gesture bursts and quickly reverts to \textit{rest}: it detects onsets but fails to sustain correct labels. This likely reflects the public datasets' statistics---short gestures with frequent returns to \textit{rest}---which degrades performance on non-rest transitions and sustained gestures.

\textbf{Window Size.} We also ablate the effect of window size, which determines the temporal context the model uses at prediction time. We vary the window from 100 timesteps (0.5 s) to 600 timesteps (3 s) to assess its impact on intent detection. \autoref{tab:window_ablation} reports results for three window sizes on the EMG-EPN-612 6-class dataset, evaluated with no lookahead to eliminate smoothing and enable direct comparison of raw performance. Smaller windows yield less stable predictions, likely because the model lacks sufficient information to distinguish transient muscle activity. Based on these results, we adopt a 600-timestep window in the final model.

\begin{table}[h]
  \centering
  \caption{Window Size Ablation - EMG-EPN-612 6-class, No Smoothing}
  \label{tab:window_ablation}
  \begin{tabular}{lcc}
    \toprule
    \textbf{Context Window Size} & \textbf{Raw Acc.} & \textbf{Transition Acc.} \\
    \midrule
    100 timesteps (0.50s) & 0.89 & 0.55\\
    300 timesteps (1.50s) & 0.90 & 0.64 \\
    600 timesteps (3.00s) & \textbf{0.92} & \textbf{0.73} \\
    \bottomrule
  \end{tabular}
\end{table}

\textbf{Buffer Size.} While we believe that our choice of buffer size is grounded in physiological factors, we also quantify how stricter tolerances affect reported transition accuracy. \autoref{tab:buffer_ablation} summarizes results as we reduce the total buffer from 200 timesteps ($\pm 500$ ms) to 120 ($\pm 300$ ms) and 60 ($\pm 150$ ms). As with the window-size ablation, we evaluate on the 6-class EMG-EPN-612 dataset in a no-lookahead setting. Tightening the tolerance yields a sharp drop in transition accuracy, highlighting the metric's flexibility and sensitivity.

\begin{table}[h]
  \centering
  \caption{Effect of Reaction Buffer Size - EMG-EPN-612 6-class, No Smoothing}
  \label{tab:buffer_ablation}
  \begin{tabular}{lcc}
    \toprule
    \textbf{Reaction Buffer Size} & \textbf{Raw Acc.} & \textbf{Transition Acc.} \\
    \midrule
    60 timesteps ($\pm$0.15s) & 0.92 & 0.52 \\
    120 timesteps ($\pm$0.30s) & 0.92 & 0.66 \\
    200 timesteps ($\pm$0.50s) & \textbf{0.92} & \textbf{0.73} \\
    \bottomrule
  \end{tabular}
\end{table}


\section{Limitations}
\label{sec:limitations}
Our approach relies on a sparse, commercial eight-channel Myo armband positioned around the forearm, which concentrates on major extensor and flexor groups but may neglect smaller muscle groups and intrinsic hand muscles. Because of this placement, it can be challenging to distinguish between certain gestures that share overlapping muscle activations. For example, wrist extension naturally activates the same extensor group involved in opening the hand, leading to potential ambiguity. To better capture these nuanced differences, additional sensors or more sophisticated electrode arrangements would be necessary. Nonetheless, the Myo armband remains an accessible choice due to its commercial availability and the breadth of existing datasets, which facilitate large-scale training and comparative benchmarking.

Although our transformer-based architecture improves on current state-of-the-art results, it requires substantially more computational resources than traditional models like SVMs or LDA. Even at a reduced inference frequency, preserving low latency necessitates running the model on a GPU, which may be impractical for wearable devices or other edge-computing scenarios. Techniques such as knowledge distillation could potentially alleviate these demands by creating lighter models suited to embedded devices. However, implementing these techniques introduces additional engineering complexities and trade-offs that are particularly challenging in resource-constrained environments.


\section{Conclusion}
\label{sec:conclusion}
We present ReactEMG, a zero-shot, real-time intent detection model that formulates sEMG intent recognition as a streaming segmentation problem. The core idea is a masked modeling objective that jointly encodes EMG and intent tokens in a single transformer, producing per-timestep predictions that react quickly at gesture onset and remain stable during holds. With a short, bounded look-ahead for aggregation, the system enables continuous, low-latency control well suited to robotic and prosthetic devices.

To enable zero-shot deployment, we pretrain on public EMG datasets and fine-tune on a posture- and task-diverse dataset that includes non-\emph{relax} transitions. We also introduce \emph{transition accuracy}, a metric that couples reaction and maintenance and better reflects online behavior than window or frame accuracy alone. Across EMG-EPN-612 (3- and 6-class) and our dataset, ReactEMG achieves state-of-the-art performance on both raw and transition accuracy. It also reduces prediction offset relative to strong sequence and convolutional baselines, all without subject-specific calibration.

By aligning muscle activity with intent via masked modeling and evaluating with a metric matched to online use, ReactEMG advances sEMG interfaces toward reliable, plug-and-play control for wearable robotics, assistive devices, and prosthetics. We release code, models, and datasets to support reuse and further study.


\section*{Acknowledgments}

This work was supported in part by an Amazon Research Award and the Columbia University Data Science Institute Seed Program. Ava Chen was supported by NIH grant 1F31HD111301-01A1. The views and conclusions contained herein are those of the authors and should not be interpreted as necessarily representing the official policies, either expressed or implied, of the sponsors. We would like to thank Katelyn Lee, Eugene Sohn, Do-Gon Kim, and Dilara Baysal for their assistance with the hand orthosis hardware. We thank Zhanpeng He and Gagan Khandate for their helpful feedback and insightful discussions.

\clearpage

\section*{REFERENCES}

\bibliographystyle{IEEEtran}
\bibliography{reactemg}

\end{document}